\documentclass[runningheads]{llncs}
\usepackage[T1]{fontenc}
\usepackage{graphicx}
\usepackage{booktabs}  
\usepackage{caption,subcaption}
\usepackage{multirow}
\usepackage[table]{xcolor}
\usepackage{url}
\usepackage{color}
\usepackage{cite}
\usepackage{graphicx}
\usepackage{booktabs}
\usepackage{pbox}
\usepackage{caption}
\usepackage{setspace}

\begin{document}
\title{Dynamics of Affective States During Takeover Requests in Conditionally Automated Driving Among Older Adults with and without Cognitive Impairment}
\titlerunning{Dynamics of Affective States During Takeover Request}

\author{
Gelareh Hajian\inst{1,2}, Ali Abedi\inst{1,2}, Bing Ye\inst{1,3}, Jennifer Campos\inst{1,4}, and Alex Mihailidis\inst{1,2,3}} 

\institute{KITE Research Institute, Toronto Rehabilitation Institute, University Health Network, Toronto, Canada 
\and
Department of Biomedical Engineering, University of Toronto
\and
Department of Occupational Science and Occupational Therapy, University of Toronto
\and
Department of Psychology, University of Toronto
}

\maketitle              
\begin{abstract}
Driving is a key component of independence and quality of life for older adults. However, cognitive decline associated with conditions such as mild cognitive impairment and dementia can compromise driving safety and often lead to premature driving cessation. Conditionally automated vehicles, which require drivers to take over control when automation reaches its operational limits, offer a potential assistive solution. However, their effectiveness depends on the driver’s ability to respond to takeover requests (TORs) in a timely and appropriate manner. Understanding emotional responses during TORs can provide insight into drivers’ engagement, stress levels, and readiness to resume control, particularly in cognitively vulnerable populations. This study investigated affective responses, measured via facial expression analysis of valence (emotional tone) and arousal (emotional intensity), during TORs among cognitively healthy older adults and those with cognitive impairment. Facial affect data were analyzed across different road geometries (straight vs. curved) and speeds (50 km/h vs. 100 km/h) to evaluate within- and between-group differences in affective states. Within-group comparisons using the Wilcoxon signed-rank test revealed significant changes in valence and arousal during TORs for both groups. Cognitively healthy individuals showed adaptive increases in arousal under higher-demand conditions, while those with cognitive impairment exhibited reduced arousal and more positive valence in several scenarios. Between-group comparisons using the Mann–Whitney U test indicated that cognitively impaired individuals displayed lower arousal and higher valence than controls across different TOR conditions. These patterns may suggest blunted emotional activation and potentially diminished situational awareness in cognitively impaired drivers. These results also highlight the potential need for adaptive, conditionally automated vehicle systems that can detect affective states and deliver supportive handover strategies tailored to cognitively vulnerable populations.
\end{abstract}


\section{Introduction}
Driving is a critical enabler of independence, mobility, and social participation among older adults. It is closely associated with quality of life, as it facilitates access to healthcare, social connections, and community engagement \cite{maresova2023challenges,qin2020driving}. However, age-related cognitive decline, particularly due to neurodegenerative conditions such as mild cognitive impairment (MCI) and dementia, can compromise driving safety and often lead to premature driving cessation \cite{teasdale2016older}. This, in turn, has been associated with higher rates of depression and increased three-year mortality \cite{fonda2001, chihuri2016}, faster cognitive decline \cite{choi2014}, lower quality of life, and disruptions to independence and personal identity \cite{sanford2018}.

Given the serious consequences of driving cessation for older adults, it is essential to explore transportation alternatives that preserve safe mobility for as long as possible \cite{chee2017, sanford2018}. Automated vehicles (AVs) offer a promising solution by reducing the cognitive demands of driving and potentially extending driving ability for individuals with cognitive impairment \cite{shergold2016, sanford2018, jang2007, hajian2024cav}. Among these, conditionally automated vehicles (CAVs), classified as Level 3 by the Society of Automotive Engineers \cite{sae2018}, enable shared control between the human driver and the automated system. In CAVs, the vehicle manages driving under specific conditions, but the human driver must resume control when a takeover request (TOR) is issued, typically in response to complex or unanticipated scenarios. While CAVs hold promise for prolonging the driving years of older adults, their safety depends on the driver’s capacity to re-engage with the driving task effectively during TORs. This challenge is particularly pronounced for older adults with cognitive impairments, who may experience delays in perception, judgment, and motor responses, placing them at elevated risk during these transitions.


Beyond behavioral and physiological indicators of takeover performance, emotional and affective responses may also play a critical role in shaping a driver’s readiness and ability to resume control \cite{du2020examining, stephenson2020effects, wang2023studying, du2020psychophysiological}. Changes in facial expressions, particularly in terms of valence (the positive or negative emotional tone) and arousal (emotional intensity), can provide insights into a driver’s cognitive-affective state during TORs \cite{stephenson2020effects, wang2023studying, du2020psychophysiological}. Despite the potential utility of these affective indicators, little is known about how such responses manifest in older adults with and without cognitive impairment during takeover events in automated vehicles.

In this study, facial expressions at the moment of TORs were analyzed across various driving scenarios characterized by differences in road geometry (straight vs. curved) and road speed (50 km/h vs. 100 km/h). Changes in facial affect, specifically valence and arousal, were examined from the period before to during TOR within each group of cognitively healthy older adults and those with cognitive impairment. The aim was to determine whether, and to what extent, changes in affective expression occurred during these safety-critical moments. This within-group analysis was used to reveal how each population emotionally responded to TORs under varying driving conditions, providing a foundation for understanding the potential role of affect in takeover moments. Additionally, a comparison analysis was conducted between the control group (older adults with normal cognition) and those with cognitive impairment during TORs to explore group-level differences in affective responses.

This work represents an initial step in a broader research effort to understand how emotional responses during takeover moments relate to driving performance and physiological markers in older adults, both with and without cognitive impairment. Ultimately, these insights could inform the design of adaptive CAV systems that more effectively support older drivers, particularly those experiencing cognitive decline. The following sections present the literature review, methodology, and results of the facial affect analysis, highlighting key differences within and between groups and across driving conditions during TORs, followed by the conclusion.

\section{Literature Review}
\label{sec:literature_review}
This section reviews existing literature on analyzing the dynamics of affective states during automated driving, focusing specifically on how emotional factors influence drivers' responses to takeover requests in CAVs.

Du et al. \cite{du2020examining} investigated how valence and arousal impact drivers' performance during TORs in Level 3 automated driving scenarios. In a simulated driving environment, 32 healthy university students (young adults, average age 21.4 years, SD = 2.9) were exposed to emotionally charged movie clips designed to induce different emotional states across the valence-arousal spectrum. Participants then experienced unexpected TORs. Takeover time and quality, measured through acceleration, jerk, and time-to-collision, were subsequently assessed. Findings indicated that positive valence significantly improved takeover quality, resulting in smoother and safer transitions (lower acceleration and jerk). However, higher arousal did not improve takeover response times, suggesting that, unlike in manual driving where heightened arousal is often linked to quicker reactions, emotional arousal had less impact in the automated driving context.

Stephenson et al. \cite{stephenson2020effects} examined arousal and eye gaze behaviours in 37 older adults (mean age = 68.4 years) during simulated Level 5 autonomous driving using the Wizard of Oz method. Electrodermal activity and heart rate were measured using Empatica E4 wristbands, while eye gaze was tracked using Tobii Pro Glasses 2. Participants experienced both expected and unexpected stops, with the latter simulating safety-critical events. Results indicated increased arousal following unexpected stops. Participants showed longer fixation durations and more frequent fixations toward the central environment (hazard location) during these events, whereas attention shifted more toward the vehicle’s human-machine interface during expected stops. This pattern suggests that unexpected events narrow attentional focus and heighten physiological responses. Although the study did not directly examine takeover requests, the unexpected stops served as analogous safety-critical events, likely triggering similar cognitive and affective responses. The findings indicate that heightened arousal and narrowed attention, as observed, are relevant factors influencing takeover performance in TOR scenarios.

Wang et al. \cite{wang2023studying} analyzed the relationship between driver arousal and takeover performance during automated driving using a high-fidelity driving simulator. A sample of 42 participants across three age groups (young, middle-aged, and older adults) engaged in simulated Level 3 autonomous driving tasks involving TORs under foggy conditions. Arousal was measured using gaze duration and pupil size, recorded via eye-tracking glasses. The results showed that, following TORs, drivers exhibited shorter gaze durations and larger pupil sizes, which were interpreted as indicators of heightened perceptual arousal and improved road information processing.

Du et al. \cite{du2020psychophysiological} investigated psychophysiological responses to TORs among 102 university students (mean age = 22.9 years) using a high-fidelity driving simulator for SAE Level 3 automated driving. Participants experienced eight TOR scenarios under varying conditions, including different cognitive loads, traffic densities, and TOR lead times. Psychophysiological measures included heart rate, galvanic skin response, gaze behaviours, and facial expressions. Emotional valence and engagement were extracted from facial expressions using the iMotions Affectiva module. The results showed that higher emotional arousal and lower emotional valence (i.e., more negative emotions) were associated with shorter TOR lead times and increased stress. Blink suppression and heart rate acceleration patterns further reflected heightened attention and negative emotional responses in urgent situations. Additionally, higher cognitive load was associated with reduced heart rate variability and narrower gaze dispersion during automated driving, indicating increased mental workload. During takeover, heart rate acceleration occurred more frequently in heavy traffic conditions, suggesting greater attentional demand and stress in more complex environments.

Huang et al. \cite{huang2024differences} examined how emotional instability affects physiological responses and takeover performance in conditionally automated driving. Forty-two healthy university students (mean age = 24.4 years) completed simulated takeover tasks under both neutral and negative emotional states, which were induced using movie clips. Emotional instability was assessed through subjective personality questionnaires and objective physiological signals, including EEG and ECG, collected via the BIOPAC MP150 system. Valence and arousal were measured using the Self-Assessment Manikin. Participants with higher emotional instability showed greater emotional reactivity, characterized by lower valence, higher arousal, elevated heart rate variability, and increased EEG power—indicators of stress, cognitive load, and reduced attention. These participants also demonstrated poorer lateral takeover performance, with higher lateral velocity, acceleration, and turning angles.

Palomares et al. \cite{palomares2023detection} investigated the detection of occupants' emotional states in AVs using physiological signals. Fifty adult drivers (aged 25–55) experienced six Level 4 automated driving scenarios, ranging from smooth driving to critical events such as system failure, within a driving simulator. ECG and EMG signals were collected, and a probabilistic emotional model was applied to estimate valence and arousal in real time. Emotional self-reports using the Self-Assessment Manikin scale confirmed the model’s predictions. For instance, the system failure scenario elicited the highest arousal and most negative valence, while the comfort scenario was associated with the lowest arousal and most positive valence.

The reviewed literature demonstrates a growing interest in understanding emotional and physiological responses during conditionally and fully automated driving, with a focus on how affective states influence or reflect takeover performance, trust, and user experience. Prior studies have examined affective dynamics using physiological signals such as heart rate variability, galvanic skin response, EEG, eye tracking, facial expression analysis, and self-reported emotion scales. While these approaches offer valuable insights, most studies have relied on healthy, young adult participants, limiting their relevance to underrepresented, at-risk populations such as older adults and individuals with cognitive impairment. Notably, none of the reviewed work explicitly targets older adults with mild cognitive impairment, who may face unique cognitive and emotional challenges during high-stakes interactions such as TORs. This is especially important given that emotional expressivity can be diminished in persons with dementia, potentially reducing the reliability of affective cues during takeover scenarios. Additionally, many existing studies rely on wearable sensors or intrusive data collection tools that may be impractical or uncomfortable for long-term, real-world use. In contrast, our work addresses this gap by focusing on older adults with mild cognitive impairment and analyzing their affective states during TORs using non-intrusive, video-based data. This approach provides a scalable and unobtrusive solution for affect monitoring in automated driving, enabling real-time adaptation of the driving experience—for example, by adjusting the timing or modality of TOR alerts or activating additional support mechanisms based on the driver’s emotional and cognitive state. Such personalization is particularly valuable for this population, who may have reduced capacity to cope with abrupt or stressful transitions.

\section{Methodology}
\subsection{Study Setting}
\begin{figure}[ht]
    \centering
    \begin{subfigure}[b]{0.7\textwidth}
        \includegraphics[width=\textwidth]{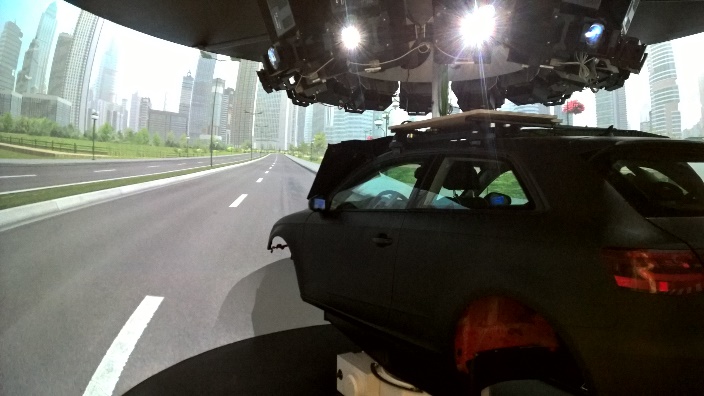}
        \caption{}
        \label{DL_fig_a}
    \end{subfigure}
    
    \vspace{0.5em} 
    
    \begin{subfigure}[b]{0.7\textwidth}
        \includegraphics[width=\textwidth]{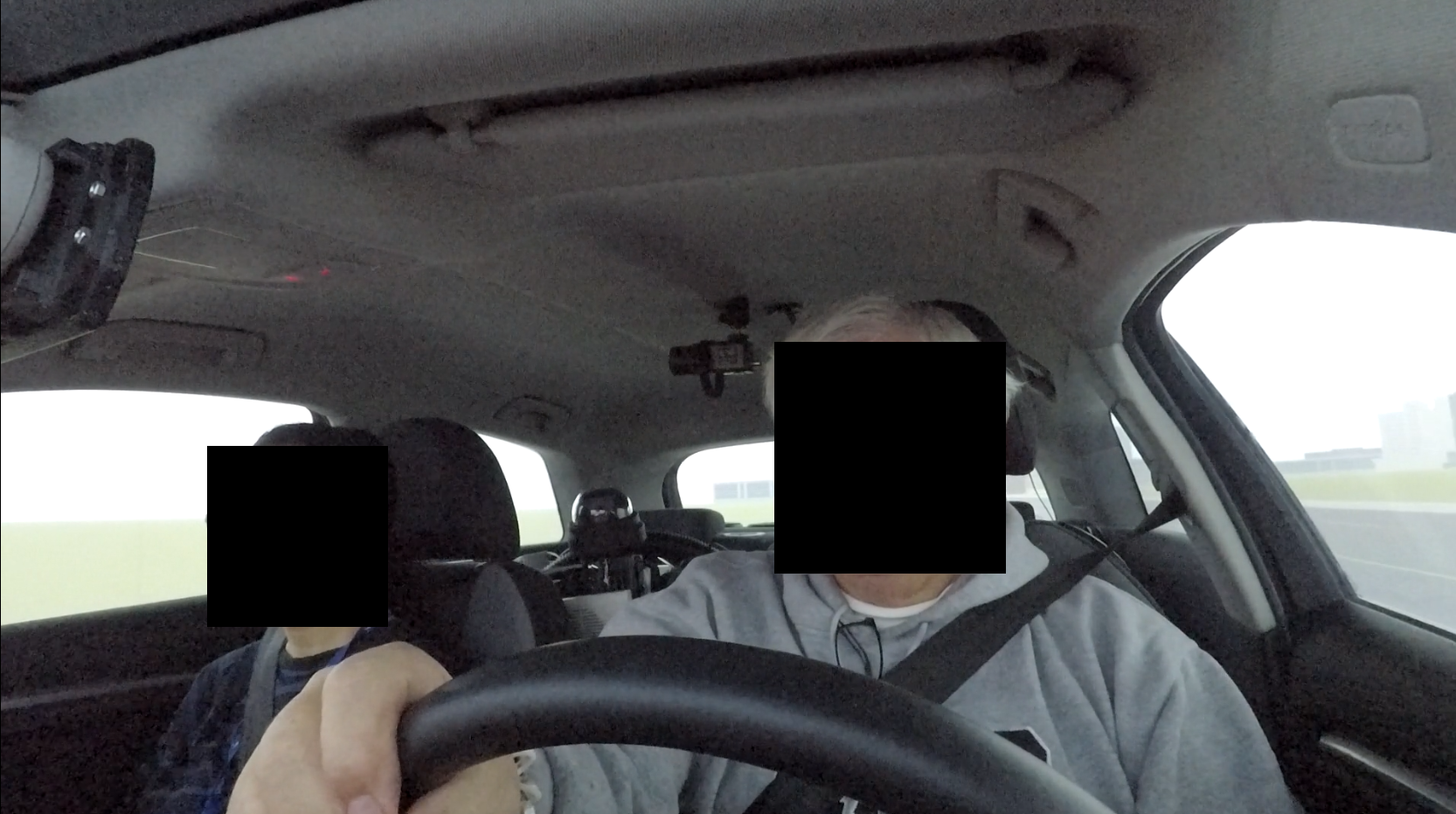}
        \caption{}
        \label{DL_fig_b}
    \end{subfigure}
    
    \caption{(a) DriverLab simulator at the Toronto Rehabilitation Institute, featuring a full-scale vehicle mounted on a motion platform within a 360-degree immersive projection dome. (b) An older adult participant engaged in an automated driving scenario in DriverLab, with a research assistant seated in the passenger seat.}

    \label{DL_fig_column}
\end{figure}

This study was conducted using a state-of-the-art driving simulator at the Toronto Rehabilitation Institute, University Health Network, in Toronto, Canada. DriverLab, the most advanced driving simulator in Canada and among the most sophisticated globally, features a full-scale passenger vehicle mounted on a turntable, which can be connected to a six-degree-of-freedom hydraulic motion platform. The vehicle retains its original controls and is equipped with customizable interfaces and integrated measurement tools. A high-resolution projection system displays onto a curved dome surrounding the vehicle, creating a fully immersive and seamless 360-degree field of view, as shown in Fig. \ref{DL_fig_a}. Figure \ref{DL_fig_b} shows an older adult participant engaged in an automated driving scenario inside the simulator.

\subsubsection{Study Design}
The study followed a 2 (cognitive status) × 2 (road geometry) × 2 (speed limit) × 2 (lighting condition) factorial design. The between-subjects factor was cognitive status, consisting of cognitively healthy individuals (control, $n = 18$) and individuals with cognitive impairment, including those with mild cognitive impairment (MCI) and very mild dementia ($n = 5$). The within-subjects factors were road geometry (straight vs. curved) and speed limit (50 km/h vs. 100 km/h). Although both daytime and nighttime scenarios were included in the experimental design, only data from daytime conditions were analyzed here due to reduced visibility in nighttime scenarios.

\subsubsection{Study Participants}
To be included in the healthy control group, participants were required to be 65 years or older, have a valid driver’s license, self-report normal motor and sensory functioning, and score $\geq 26$ on the Montreal Cognitive Assessment (MoCA) \cite{Moca}. For inclusion in the cognitively impaired group, participants with mild cognitive impairment (MCI) or very mild dementia had to be 65 years or older, self-report a clinical diagnosis of MCI or very mild dementia from a healthcare provider, report no motor or sensory impairments, score 0.5 on the Clinical Dementia Rating Scale (CDR) \cite{CDR}, and either hold a valid driver’s license or have voluntarily stopped driving within the past 18 months. One participant in the cognitively impaired group did not complete the CDR assessment due to the absence of an available informant, which is required for CDR administration. In this case, group assignment was based on the participant’s clinical diagnosis provided by their physician.

\subsubsection{Study Procedures}
The study involved two visits to DriverLab. The first visit included a simulator adaptation session and the administration of screening and baseline assessments. Screening procedures involved obtaining informed consent and completing the Health History Questionnaire, which covered age, medication use, and general health status. Baseline assessments evaluated cognitive and functional abilities across key domains, including attention, memory, executive function, and processing speed. During this visit, participants also completed a 10-minute adaptation drive to familiarize themselves with the simulator environment and Level 3 automation, while reducing the likelihood of simulator sickness. This session included a practice takeover task to ensure participants were comfortable and capable of responding to takeover requests. Participants were allowed to repeat the practice session if needed. Individuals who experienced significant simulator sickness during this visit did not proceed to the second visit.

The second visit included two 12-minute conditionally automated driving sessions under varying lighting (daytime or nighttime) and road conditions. In each session, the vehicle was initially operated by the automated system. At predetermined points, an audio-visual takeover request (TOR) was issued, prompting participants to resume manual control. Each session included four TOR events, covering different combinations of road geometry (straight and curved) and speed limits (50 km/h and 100 km/h). One session was conducted during the daytime and the other at nighttime; session order was counterbalanced across participants. Breaks were provided between scenarios, during which participants were screened for simulator sickness using the Fast Motion Sickness Scale (FMSS) \cite{FMSS}. Scheduled rest periods supported recovery and helped minimize simulator sickness. If mild symptoms were reported, additional breaks were offered; sessions were discontinued in cases of severe symptoms. To enhance comfort and reduce the likelihood of dropout, the environment was kept cool with steady airflow. For the purpose of this paper, only data from the daytime sessions were analyzed. Nighttime facial expression data were excluded due to unreliable visibility conditions.

The study was approved by the University Health Network Research Ethics Board (REB \#20-5090). Informed consent was obtained from all participants using plain-language forms. For individuals with mild cognitive impairment (MCI) or mild dementia, a caregiver or informant co-signed the consent form as an additional safeguard. Participants were monitored throughout the study, and sessions were paused or discontinued if necessary to ensure their well-being.

\subsection{Analysis of Affective State Dynamics}
\label{affective_state_estimation}
Video recordings of drivers during conditionally automated driving sessions were captured using a GoPro camera at a resolution of $1920 \times 1080$ pixels and 48 frames per second. The videos were downsampled to 16 frames per second using FFmpeg \cite{ffmpeg} to reduce computational load while maintaining sufficient temporal resolution for affective analysis. This frame rate was selected based on prior research indicating that 10–15 frames per second is typically adequate for capturing facial expression dynamics \cite{sariyanidi2014automatic}. The videos were then spatially cropped using FFmpeg \cite{ffmpeg} to exclude the front-seat passenger (the research assistant) and retain only the driver’s region in each frame.

Facial region detection was performed using MediaPipe Face Detection \cite{lugaresi2019mediapipe}, which generated bounding boxes around the face in each frame. These cropped face images were processed using EmoFan (Emotion Face Alignment Network) \cite{toisoul2021estimation}, a deep learning model pre-trained on AffectNet \cite{mollahosseini2017affectnet}, to compute frame-level continuous values of valence and arousal. Sixteen valence and sixteen arousal scores were extracted per second of video, with each score represented as a real-valued number ranging from $-1$ to $+1$, enabling high-resolution temporal tracking of affective states throughout the driving session.

\begin{figure}
    \centering
        \includegraphics[width=0.75\textwidth]{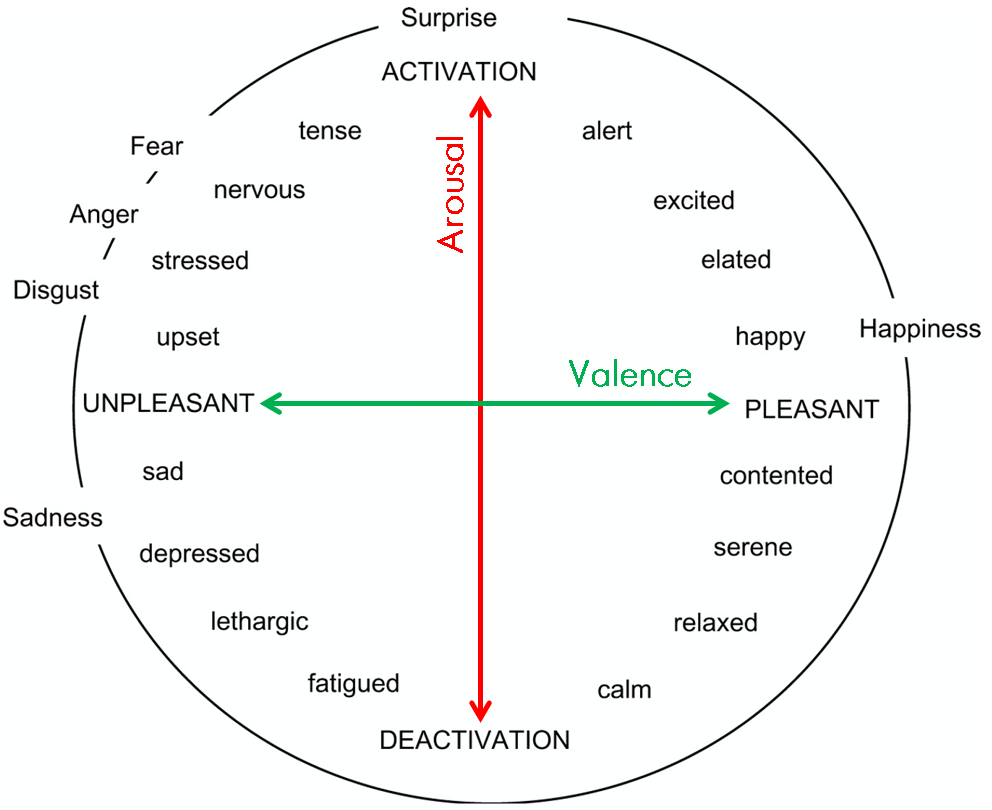}
    \caption{Russell’s circumplex model of affect \cite{russell1980circumplex}, with valence on the horizontal axis and arousal on the vertical axis.}
    \label{fig:circumplex}
\end{figure}

Valence represents the positivity or negativity of an emotional state, indicating its pleasantness or unpleasantness, while arousal reflects the intensity of the emotion, ranging from calm or deactivated to excited or activated states. Together, these two dimensions form the basis of the circumplex model of affect \cite{russell1980circumplex}, a widely used framework in affective computing (see Fig.~\ref{fig:circumplex}).

\subsection{Statistical Analysis}
\subsubsection{Within-group Analysis}
To examine the affective dynamics of drivers during TORs, the primary dependent variables, valence and arousal (derived from facial expression data), were compared between the periods before and during TOR events. The Wilcoxon signed-rank test with a two-sided hypothesis was used to assess whether affective responses during TORs differed significantly from those observed during the preceding automated driving period. This non-parametric test was chosen due to the non-normal distribution and ordinal nature of the data, making it suitable for comparing two related samples. Analyses were conducted separately for cognitively healthy older adults and those with cognitive impairment. Both \textit{p}-values and standardized mean differences (SMDs) were reported to evaluate the statistical and practical significance of the observed changes.

\subsubsection{Between-group Analysis}
To compare affective responses between independent groups, specifically, healthy older adults and those with cognitive impairment after TORs, the Mann–Whitney U test was applied using a two-sided hypothesis. This test was appropriate due to the non-normal distribution and unequal group sizes (control group: $n = 18$; cognitive impairment group: $n = 5$). The analysis evaluated whether TORs were associated with statistically significant changes in emotional responses across groups.

\section{Results and Discussion}
\begin{figure}
    \centering
        \includegraphics[width=.8\textwidth]{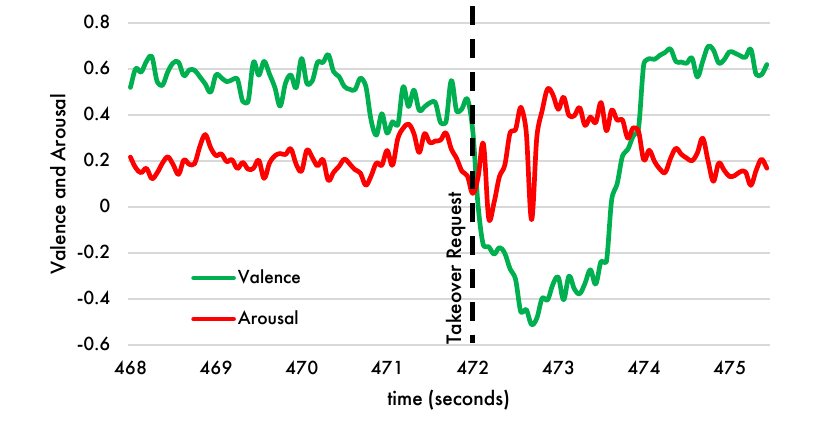}
    \caption{Changes in valence and arousal values for an older adult with mild cognitive impairment before and after a takeover request (TOR) occurring at second 472, during the high-speed (100km/h) straight condition.}
    \label{fig:affect}
\end{figure}

Facial expressions during TORs were analyzed to assess valence and arousal in older adults with and without cognitive impairment. Changes from pre-TOR to TOR were examined within each group across road geometries (straight vs. curved) and speed (low speed (50 km/h) vs. high speed (100 km/h)). The aim was to evaluate within-group affective changes and between-group differences. An example of changes in valence and arousal values for an older adult pre- and during a TOR is shown in Fig. \ref{fig:affect}.

\subsection{Facial Affect Analysis Results}
\subsubsection{Statistical Analysis Results -- Within-group Analysis}
Tables~\ref{tab:control_results} (a) and (b) present within-group differences in valence and arousal before and during TORs across four road conditions: low speed (50 km/h)-straight, low speed (50 km/h)-curved, high speed (100 km/h)-straight, and high speed (100 km/h)-curved, for cognitively healthy older adults and those with cognitive impairment. For each condition, \textit{p}-values from the Wilcoxon signed-rank test and SMDs are reported. A positive SMD indicates an increase in valence or arousal from the automated driving period to the TOR moment, whereas a negative SMD indicates a decrease.

\begin{table}[ht]
\centering
\caption{Changes in (a) valence and (b) arousal during takeover requests (TORs), compared to time periods preceding TORs, across different TOR conditions in healthy older adults and those with cognitive impairment. \textit{p}-values from the Wilcoxon signed-rank test and standardized mean differences (SMDs) are reported. The expected cognitive demands for the different TOR conditions were considered as follows: low for low speed–straight, medium for both low speed–curved and high speed–straight, and high for high speed–curved.}
\label{tab:control_results}

\subfloat[Valence]{
\resizebox{\textwidth}{!}{%
\begin{tabular}{l@{\hspace{1em}}c@{\hspace{1em}}c@{\hspace{1em}}c@{\hspace{1em}}c@{\hspace{3em}}c@{\hspace{1em}}c@{\hspace{1em}}c@{\hspace{1em}}c}
\hline
 & \multicolumn{4}{c@{\hspace{3em}}}{\textbf{Cognitively Healthy}} & \multicolumn{4}{c}{\textbf{Cognitively Impaired}} \\
\textbf{TOR condition} & \textbf{Before} & \textbf{During} & \textbf{\textit{p}-value} & \textbf{SMD} & \textbf{Before} & \textbf{During} & \textbf{\textit{p}-value} & \textbf{SMD} \\
\hline
Low speed-straight     & $-0.22\pm0.11$ & $-0.16\pm0.13$ & < 0.001 & $0.43$  & $-0.11\pm0.11$ & $-0.14\pm0.10$ & < 0.001 & $-0.29$ \\
Low speed-curved       & $-0.23\pm0.10$ & $-0.18\pm0.12$ & < 0.001 & $0.33$  & $-0.13\pm0.10$ & $-0.04\pm0.20$ & < 0.001 & $0.57$  \\
High speed-straight    & $-0.21\pm0.11$ & $-0.26\pm0.16$ & < 0.001 & $-0.35$ & $-0.08\pm0.27$ & $-0.07\pm0.32$ & < 0.001 & $0.15$  \\
High speed-curved      & $-0.24\pm0.20$ & $-0.22\pm0.24$ & < 0.001 & $-0.06$ & $-0.10\pm0.12$ & $-0.24\pm0.07$ & < 0.001 & $-1.11$ \\
\hline
\end{tabular}
}
}

\vspace{1.5em}

\subfloat[Arousal]{
\resizebox{\textwidth}{!}{%
\begin{tabular}{l@{\hspace{1em}}c@{\hspace{1em}}c@{\hspace{1em}}c@{\hspace{1em}}c@{\hspace{3em}}c@{\hspace{1em}}c@{\hspace{1em}}c@{\hspace{1em}}c}
\hline
 & \multicolumn{4}{c@{\hspace{3em}}}{\textbf{Cognitively Healthy}} & \multicolumn{4}{c}{\textbf{Cognitively Impaired}} \\
\textbf{TOR condition} & \textbf{Before} & \textbf{During} & \textbf{\textit{p}-value} & \textbf{SMD} & \textbf{Before} & \textbf{During} & \textbf{\textit{p}-value} & \textbf{SMD} \\
\hline
Low speed-straight     & $0.16\pm0.11$ & $0.12\pm0.12$ & < 0.001 & $-0.30$ & $0.15\pm0.07$ & $0.12\pm0.09$ & < 0.001 & $-0.35$ \\
Low speed-curved       & $0.14\pm0.11$ & $0.17\pm0.12$ & 0.0011  & $0.25$  & $0.16\pm0.07$ & $0.11\pm0.09$ & < 0.001 & $-0.52$  \\
High speed-straight    & $0.15\pm0.11$ & $0.17\pm0.15$ & < 0.001 & $0.15$  & $0.14\pm0.07$ & $0.16\pm0.14$ & 0.0012  & $0.20$  \\
High speed-curved      & $0.15\pm0.12$ & $0.17\pm0.13$ & < 0.001 & $0.20$  & $0.14\pm0.07$ & $0.14\pm0.12$ & 0.0224  & $0.01$  \\
\hline
\end{tabular}
}
}
\end{table}

\noindent
\textbf{Valence--} Significant changes in valence were observed in all TOR conditions for both groups (\textit{p} < 0.001 for all comparisons). Among cognitively healthy older adults, the greatest increases in valence occurred in the low speed-straight (SMD = 0.43) and low speed-curved (SMD = 0.33) conditions, indicating moderate positive shifts in emotional tone. Conversely, valence decreased slightly in high speed conditions, with a small negative shift in high speed-straight (SMD = –0.35) and a minimal change in high speed-curved (SMD = –0.06), 
suggesting mildly heightened unpleasant emotion or reduced emotional positivity during high-speed TORs. For the cognitively impaired group, valence also changed significantly across all conditions, but the direction and magnitude of effects varied more substantially. 
A notable increase was observed in the low speed–curved condition (SMD = 0.57), suggesting a strong shift toward more positive emotional tone. This may reflect greater comfort in moderate-demand scenarios or a different perception of the event’s complexity—potentially indicating that participants with cognitive impairment did not perceive the event as being as risky or demanding as it objectively was. In contrast, a sharp decline in valence was found in the high speed–curved condition (SMD = –1.11), indicating a marked increase in unpleasant emotional experience during the most cognitively demanding TORs. Smaller shifts were seen in low speed–straight (SMD = –0.29) and high speed–straight (SMD = 0.15), suggesting minimal or inconsistent emotional responses in those conditions.


\noindent
\textbf{Arousal--} Arousal responses also showed statistically significant changes in most conditions, for both groups. 
Among cognitively healthy older adults, arousal decreased in the low speed–straight condition (SMD = –0.30), indicating a reduction in emotional activation. This may suggest that the event was perceived as low risk, leading to a more relaxed or disengaged state. In contrast, arousal increased in all other conditions: low speed–curved (SMD = 0.25), high speed–straight (SMD = 0.15), and high speed–curved (SMD = 0.20). These increases suggest greater attentional engagement and heightened alertness in response to more cognitively demanding scenarios, particularly those involving curves or high speeds. This pattern reflects an adaptive emotional response, with participants modulating arousal in accordance with perceived task complexity and urgency. In contrast, older adults with cognitive impairment demonstrated more variable arousal responses. Arousal significantly decreased in both low speed–straight (SMD = –0.35) and low speed–curved (SMD = –0.52) conditions, with the larger decline in the curved scenario suggesting blunted emotional engagement even in moderately demanding contexts. Arousal increased slightly in the high speed–straight condition (SMD = 0.20), indicating some degree of cognitive or emotional engagement. However, arousal remained almost unchanged in the high speed–curved condition (SMD = 0.01), despite it being the most cognitively demanding scenario. This minimal response could indicate impaired situational awareness or a failure to perceive the complexity of the driving task.



Overall, cognitively healthy participants exhibited more consistent and adaptive affective responses, characterized by increased arousal and moderately increased valence in low speed–curved conditions, along with small arousal increases in high-speed scenarios. In contrast, individuals with cognitive impairment showed greater variability in both valence and arousal. While valence increased substantially in the low speed–curved condition, it sharply decreased under high speed–curved scenarios. Arousal responses were generally blunted across conditions in the cognitively impaired group, with significant decreases even in moderately demanding contexts and minimal change in the most complex condition. These patterns suggest reduced emotional engagement or less effective modulation of affective states during TORs, particularly under higher cognitive demand.

\vspace{0.3 cm}


\subsubsection*{Statistical Analysis Results – Between-group Analysis}
Table~\ref{tab:between_results} presents the results of the between-group comparison of valence and arousal values following TORs across different driving conditions. Comparisons were made between those with cognitive impairment and cognitively healthy older adults using the Mann–Whitney U test. For each condition, \textit{p}-values and SMDs are reported.

\begin{table}[ht]
\centering
\caption{Differences in valence and arousal values between older adults with cognitive impairment and cognitively healthy older adults in time periods during takeover requests (TORs) across different TOR conditions. Reported values include p-values from the Mann–Whitney U test and standardized mean differences (SMDs).}
\label{tab:between_results}
\begin{tabular}{l@{\hspace{1.5em}}c@{\hspace{1.5em}}c@{\hspace{3.5em}}c@{\hspace{1.5em}}c}
\hline
 & \multicolumn{2}{c@{\hspace{1.5em}}}{\textbf{Valence}} & \multicolumn{2}{c}{\textbf{Arousal}} \\
\hline
\textbf{TOR condition} & \textbf{\textit{p}-value} & \textbf{SMD} & \textbf{\textit{p}-value} & \textbf{SMD} \\
\hline
Low speed-straight     & < 0.001 & 0.36  & 0.0946 & 0.13 \\
Low speed-curved       & < 0.001 & 0.91  & < 0.001 & -0.42  \\
High speed-straight   & < 0.001 & 0.86 & < 0.001 & -0.20  \\
High speed-curved     & < 0.001 & -0.07 & 0.0029 & -0.22 \\
\hline
\end{tabular}
\end{table}

\noindent
\textbf{Valence--} Significant between-group differences in valence were found across all TOR conditions (\textit{p} < 0.001). The largest differences were observed in the low speed-curved (SMD = 0.91) and high speed-straight (SMD = 0.86) conditions, where the cognitively impaired group showed higher valence values than the control group. This suggests that, in these scenarios, the cognitively impaired participants exhibited a more positive or less negative emotional tone following TORs. This may reflect less perceived threat or stress to moderately demanding events (curved and high speed cases) in cognitively impaired participants than in controls. Moderate group differences were also observed in the low speed-straight condition (SMD = 0.36), while the smallest and only slightly negative difference occurred in the high speed-curved condition (SMD = –0.07), indicating relatively similar valence levels between the groups.

\noindent
\textbf{Arousal--} Arousal differences between groups were generally smaller in magnitude but statistically significant in most conditions. In low speed-curved (SMD = –0.42), high speed-straight (SMD = –0.20), and high speed-curved (SMD = –0.22) conditions, the cognitively impaired group exhibited lower arousal levels compared to the control group, suggesting reduced emotional activation, alertness, or attentional response in this group in conditions requiring moderate to high engagement. In the low speed-straight condition, the difference was not statistically significant (\textit{p} = 0.0946, SMD = 0.13), indicating comparable levels of arousal across groups in simpler road scenarios.


These results highlight important affective differences during TORs between cognitively healthy and impaired older adults. While the cognitively impaired group often experienced more positive valence, they also showed reduced arousal in moderately and highly demanding conditions, suggesting a possible mismatch between emotional tone and task complexity. This pattern may reflect impaired risk perception, reduced emotional reactivity, or altered engagement.




\subsection{limitation}
A key limitation of this study is the small sample size for the cognitively impaired group, which may reduce statistical power, increase the risk of Type I and II errors, lead to unstable effect size estimates, and limit the generalizability of between-group comparisons. Future work should aim for larger, balanced groups (at least 15–20 participants per group) to improve reliability and generalizability.

Additionally, this study relied on facial expression data as the sole indicator of affective states. While facial expressions provide valuable insight into emotional responses, they may not fully capture the complexity of internal emotional or cognitive states, particularly in older adults with cognitive impairment, who may exhibit blunted or atypical facial expressions. Future studies should include larger and more diverse samples and consider integrating multimodal physiological measures (e.g., heart rate variability via photoplethysmography and gaze patterns via eye-tracking) to gain a more comprehensive understanding of emotional and cognitive readiness during TORs.

Moreover, facial affect estimation models are often trained on younger, neurotypical populations and may not generalize well to older adults, particularly those with neurodegeneration, whose facial expressions can be subtler, atypical, or less dynamically expressed. This introduces potential biases in affect recognition.

Finally, as data collection occurred in a controlled lab setting, real-world factors such as lighting variability, driver fatigue, and external distractions were not fully represented. These factors may influence both emotional responses and the performance of models trained under ideal conditions. Future work should validate model robustness in naturalistic driving environments to enhance generalizability.

\section{Conclusion}
This study examined affective responses, specifically valence and arousal, during TORs in CAVs among cognitively healthy older adults and those with cognitive impairment. Within-group analyses revealed that both groups exhibited significant emotional shifts during TORs, although the direction and intensity of these changes varied depending on road context. Notably, cognitively impaired participants showed increased valence in two conditions, potentially reflecting a diminished perception of risk, and lower arousal in low-speed settings, suggesting reduced emotional engagement or alertness during these safety-critical transitions. Between-group comparisons further supported these patterns, with the cognitively impaired group displaying a mismatch between emotional tone and intensity relative to their healthy counterparts.

These findings highlight important differences in how older adults with cognitive impairment emotionally experience TORs, raising concerns about their readiness to resume control in CAVs. The observed blunting of arousal, particularly in more demanding scenarios, may compromise takeover performance and highlights the need for adaptive vehicle systems that can better support this population. Future work will examine the relationship between facial expressions and takeover performance and explore how affective and physiological signals can be integrated to predict takeover readiness.

\vspace{8mm}
\noindent
\textbf{Acknowledgment--}This research was supported by the Canadian Institutes of Health Research (CIHR), AGE-WELL (Canada’s Technology and Aging Network), and the Alzheimer's Society of Canada.

\begingroup
\setstretch{1.0} 
\bibliographystyle{IEEEtran}
\bibliography{references}
\endgroup

\end{document}